\documentclass[runningheads]{llncs}

\usepackage[mobile]{eccv}

\usepackage{eccvabbrv}

\usepackage{graphicx}
\usepackage{booktabs}

\usepackage[accsupp]{axessibility}  

\usepackage[pagebackref,breaklinks,colorlinks,citecolor=eccvblue]{hyperref}

\usepackage{orcidlink}

\usepackage{tcolorbox}
\usepackage{array}
\usepackage{multirow}
\usepackage{enumitem}
\usepackage{colortbl}
\usepackage{tabularx}
\usepackage{twemojis}
\usepackage{pifont}

\newcommand{\xmark}{\ding{55}}%
\newcommand{\cmark}{\ding{51}}%
\definecolor{lightgray}{gray}{0.9}
\newcommand{\tablestyle}[2]{\setlength{\tabcolsep}{#1}\renewcommand{\arraystretch}{#2}\centering\footnotesize}

\begin{document}

\title{Groma: Localized Visual Tokenization for Grounding Multimodal Large Language Models} 

\titlerunning{Groma: Localized Visual Tokenization for MLLMs}

\author{
Chuofan Ma$^{1}$\thanks{Work done during Chuofan's internship at ByteDance.}~~~
Yi Jiang$^{2\dagger}$~~~
Jiannan Wu$^{1}$~~~
Zehuan Yuan$^{2}$~~~
Xiaojuan Qi$^{1}$\thanks{Corresponding authors}~~~
}

\authorrunning{Ma et al., 2024}

\institute{$^1$The University of Hong Kong~~~ $^2$ByteDance Inc.~~~}

\maketitle

\begin{abstract}
  We introduce Groma, a Multimodal Large Language Model (MLLM) with grounded and fine-grained visual perception ability. Beyond holistic image understanding, Groma is adept at region-level tasks such as region captioning and visual grounding. Such capabilities are built upon a localized visual tokenization mechanism, where an image input is decomposed into regions of interest and subsequently encoded into region tokens. By integrating region tokens into user instructions and model responses, we seamlessly enable Groma to understand user-specified region inputs and ground its textual output to images. Besides, to enhance the grounded chat ability of Groma, we curate a visually grounded instruction dataset by leveraging the powerful GPT-4V and visual prompting techniques. Compared with MLLMs that rely on the language model or external module for localization, Groma consistently demonstrates superior performances in standard referring and grounding benchmarks, highlighting the advantages of embedding localization into image tokenization. Project page: \href{https://groma-mllm.github.io/}{https://groma-mllm.github.io/}.
\end{abstract}

\vspace{-8mm}
\begin{figure}[h]
\centering
\includegraphics[width=0.95 \textwidth]{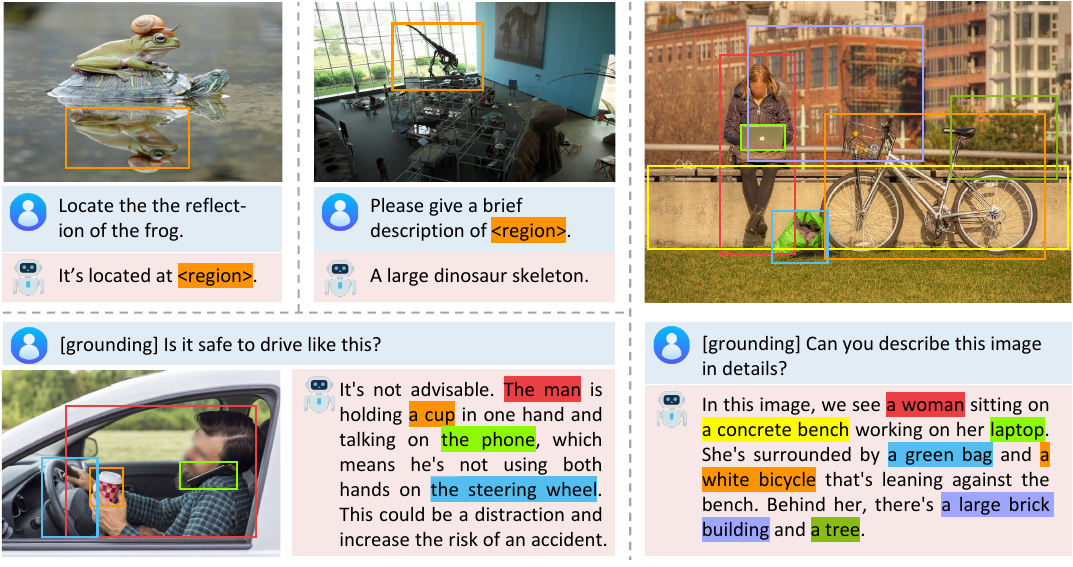}
\vspace{-2mm}
\caption{Groma is a multimodal large language model with exceptional region understanding and visual grounding capabilities. It can take user-defined region inputs (boxes) as well as generate long-form responses that are grounded to visual context.}
\label{fig:teaser}
\vspace{-4mm}
\end{figure}

\section{Introduction}
\label{sec:intro}


Multimodal Large Language Models (MLLMs) have spread the sparks of artificial general intelligence~\cite{bubeck2023sparks} from language to the visual domain~\cite{liu2024visual, zhu2023minigpt, dai2023instructblip, ye2023mplug, wang2023cogvlm}. Owing to the foundational capabilities of Large Language Models (LLMs)~\cite{chatgpt, gpt4, touvron2023llama, touvron2023llama2, chiang2023vicuna}, MLLMs excel in vision-language tasks that require advanced understanding and reasoning, such as image captioning and visual question answering. However, despite these achievements, current MLLMs typically fall short of localization capabilities, thus cannot ground understanding to the visual context. Such limitations constrains the model from fulfilling its potential in real-world applications like robotics, autonomous driving, and augmented reality.

\begin{figure}[h]
\centering
\includegraphics[width=1.0 \textwidth]{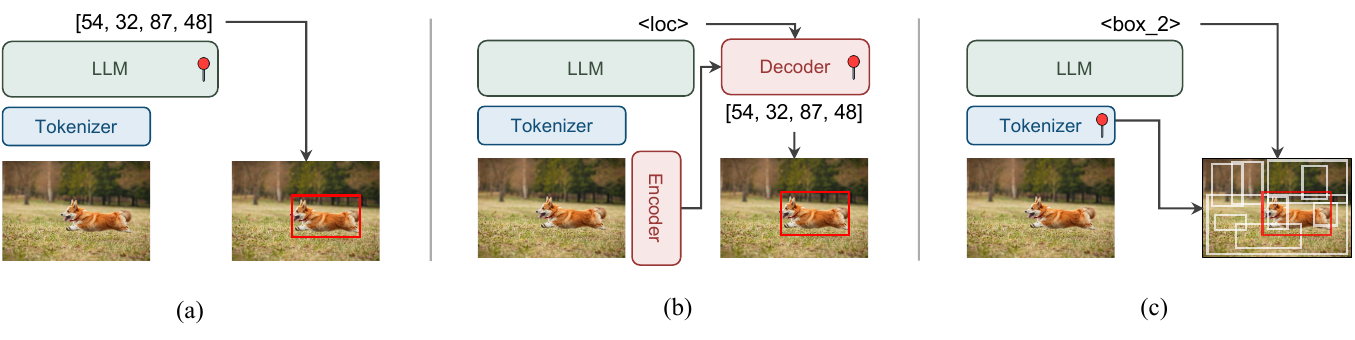}
\caption{Different paradigms of grounded MLLMs. We mark the modules for localization with \twemoji{round pushpin}. (a) LLM for localization (\eg, Kosmos-2~\cite{peng2023kosmos} and Shikra~\cite{chen2023shikra}); (b) External modules for localization (\eg, Lisa~\cite{lai2023lisa}); and (c) Localized visual tokenization (Ours).}
\label{fig:paradigm}
\end{figure}

In light of the gap, one stream of research attempts to augment the LLM to directly output quantized object coordinates for localization~\cite{peng2023kosmos, chen2023shikra, you2023ferret, chen2023minigpt, bai2023qwen-vl, wang2023cogvlm} (\cref{fig:paradigm}\textcolor{red}{(a)}). While this method is simple in design, the substantial computational demands of LLMs make it challenging to process high-resolution image inputs, which are essential for accurate localization. Besides, the nature of sequence outputs in LLMs is not well-suited for dense prediction tasks such as segmentation. These concerns elicit another stream of research, which incorporates an external localization module (\eg, SAM~\cite{kirillov2023sam}) to decode bounding boxes or masks~\cite{lai2023lisa, rasheed2023glamm, zhang2023llava, pi2023detgpt} (\cref{fig:paradigm}\textcolor{red}{(b)}). This approach circumvents aforementioned issues, but introduces additional latency in inference as it requires processing the image input twice with the MLLM and the localization module, respectively. 

The above motivates us to explore a new paradigm for grounded MLLMs. Drawing inspiration from open-vocabulary object detection~\cite{zhou2022detecting}, we decompose the grounding task into two sub-problems: discovering the object (localization) and relating the object to texts (recognition). We notice that localization alone requires little semantic understanding but demands perceptual skills, which is typically out of the scope of an LLM's expertise. This inspires us to decouple localization and recognition within MLLMs. But instead of using external modules, we propose exploiting the spatial understanding capability in the visual tokenizer of MLLMs for localization (\cref{fig:paradigm}\textcolor{red}{(c)}). This perceive-then-understand design also resembles human vision process.  

Building upon this concept, we introduce Groma\footnote{In Latin, Groma refers to an instrument used for accurate measurement, which implies our focus on accurate localization for MLLMs.} (\textbf{Gro}unded \textbf{M}ultimodal \textbf{A}ssistant), an MLLM with localized and fine-grained visual perception abilities. Specifically, Groma incorporates region tokenization alongside standard image tokenization to identify and encode potential regions of interest (ROIs) into region tokens. During this process, location information is extracted from the image and associated with region tokens, with each region token anchored to the underlying ROI. This allows Groma to ground its textual output by simply referring to region tokens, alleviating the need for the LLM to meticulously regress object coordinates. Moreover, the tokenizer of Groma can also encode user-specified region inputs (\ie, bounding boxes) into region tokens, which are directly inserted into user instructions to initiate referential dialogue.

Compared to previous methods that augment LLMs for localization~\cite{peng2023kosmos, chen2023shikra, you2023ferret, chen2023minigpt}, Groma circumvents the heavy computation of LLMs when handling high-resolution input by settling localization to the image tokenization process. That is, Groma can use high-resolution images for tokenizer input and downsampled image tokens for LLM input, which saves computation without sacrificing localization accuracy. Besides, unlike methods adopting separate designs for modeling grounding outputs and referring inputs~\cite{rasheed2023glamm, zhang2023llava}, Groma seamlessly unifies the two capabilities with the use of region tokens.

From the data perspective, to improve the localized understanding of Groma, we adopt an extensive collection of datasets with region-level annotations for training, which encompasses a range of region semantics from objects and relationships to detailed region descriptions. In addition, to remedy the lack of long-form grounded data, we construct a visually grounded chat dataset called Groma Instruct for instruction finetuning. Groma Instruct is the first grounded chat dataset constructed with both visual and textual prompts, leveraging the powerful GPT-4V for data generation.

Our comprehensive experiments demonstrate the superiority of the design of Groma, with results showing that it outperforms all comparable MLLMs on established referring and grounding benchmarks. We also showcase that Groma maintains strong image-level understanding and reasoning abilities on the conversational VQA benchmark. Moreover, to assess the ability to localize multiple, diverse, and variably-sized objects, we adapt the LVIS~\cite{gupta2019lvis} detection benchmark for object grounding evaluation. On this challenging benchmark, Groma surpasses alternative methods by a significant margin (over $10\%$ AR), highlighting its robust and precise localization capabilities.

\vspace{-4mm}

\section{Related Work}
\subsubsection{Image-level MLLMs.} Large language models (LLMs) such as GPT series~\cite{wei2022gpt-3, achiam2023gpt-4} and LLaMA\cite{touvron2023llama, touvron2023llama2} have recently undergone rapid development and sparked a revolution in the field of natural language processing. Such progress inspires the community to extend the foundational capabilities of LLMs to the visual domain, giving birth to multimodal large language models (MLLMs). The pioneering works~\cite{li2023blip-2, alayrac2022flamingo, zhang2023llava, zhu2023minigpt, dai2023instructblip, ye2023mplug, li2023otter} of MLLMs typically follow a tripartite architecture, comprising a visual encoder, a vision-language connector, and a large language model. Specifically, BLIP-2~\cite{li2023blip-2} and Flamingo~\cite{alayrac2022flamingo} first propose the Q-Former/Resampler to bridge vision and language. LLaVA~\cite{zhang2023llava} and MiniGPT4~\cite{zhu2023minigpt} streamline this vision-language connector to a linear layer, and introduce visual instruction tuning to enhance the instruction-following ability of MLLMs. Following works~\cite{chen2023internvl, wang2023cogvlm} further showcase the immense potential of MLLMs by scaling up the visual components to the magnitude as LLMs. While these works have exhibited impressive visual understanding capabilities, they are predominantly constrained to image-level tasks, such as image captioning and image visual question answering. This necessitates the research into region-level MLLMs, which unlock more nuanced and granular visual-language interactions.

\subsubsection{Region-level MLLMs.} In pursuit of fine-grained and grounded image understanding, recent studies further integrate region-level data into the training of MLLMs~\cite{wang2024visionllm, peng2023kosmos, chen2023shikra, wang2023all-seeing, chen2023minigpt, zhao2023bubogpt, yuan2023osprey}. In particular, to model box inputs and outputs, Kosmos-2~\cite{peng2023kosmos} and Shikra~\cite{chen2023shikra} directly quantize bounding boxes into discrete location tokens or numeric representation of positions. GPT4RoI~\cite{zhang2023gpt4roi} and RegionGPT~\cite{guo2024regiongpt} use a simple pooling operation to extract the features within boxes or masks as the region representations. While Ferret~\cite{you2023ferret} proposes a spatial-aware visual sampler to deal with free-form region inputs. Besides, to achieve more accurate localization, some works~\cite{lai2023lisa, rasheed2023glamm, zhang2023llava-grounding} resort to off-the-shelf models for pixel-level grounding. For instance, LISA~\cite{lai2023lisa} takes the segmentation token generated by the MLLM as the prompts for SAM~\cite{kirillov2023sam} to produce the segmentation masks. GLaMM~\cite{rasheed2023glamm} and LLaVA-Ground~\cite{zhang2023llava} further advance the concept and enable grounded conversation generation. Our work shares the same focus with the aforementioned methods on region-level understanding and grounding. Yet, we distinguish ourselves from existing studies by proposing a novel perspective in enhancing the localization ability of MLLMs.

\section{Method}
In this section, we present Groma, a grounded multimodal large language model capable of understanding user-defined region inputs and generating visually grounded outputs. We first illustrate the model architecture of Groma in \cref{sec:arch}. Then we introduce how to format region input and output in \cref{sec:format}. Finally, we detail the learning pipelines \cref{sec:train}.

\begin{figure}[h]
\centering
\includegraphics[width=1.0 \textwidth]{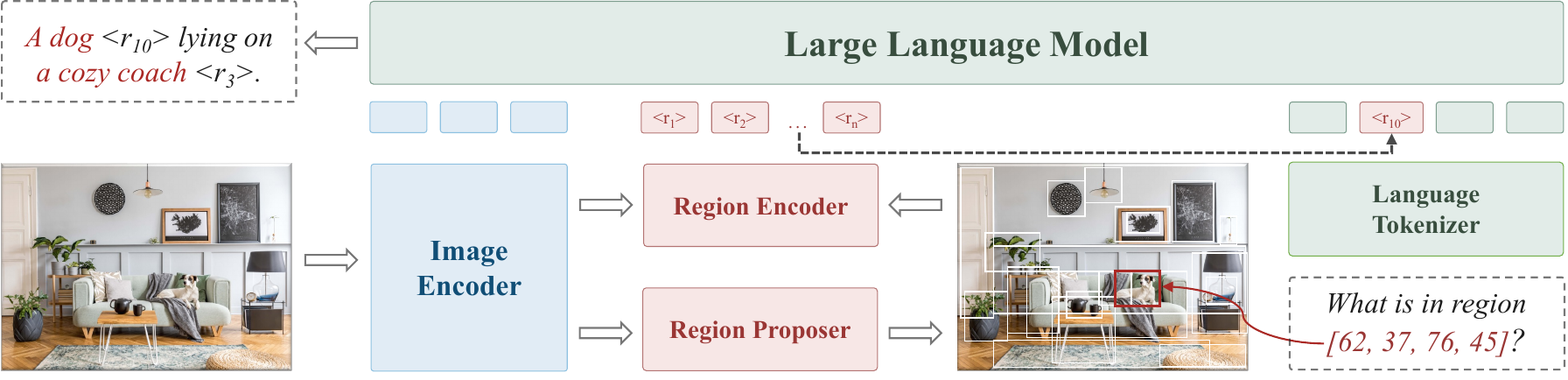}
\caption{\textbf{Overview of Groma}. Groma encodes the image input into both global image tokens and local region tokens. For region tokenization, a general-purpose region proposer is introduced to discover regions of interest, followed by a light-weight region encoder. By integrating region tokens into user instructions and model responses, Groma unlocks the referring and grounding abilities of MLLMs.}
\label{fig:arch}
\vspace{-3mm}
\end{figure}

\subsection{Model Architecture}
\label{sec:arch}
As illustrated in \cref{fig:arch}, Groma primarily consists of (1) an image encoder for scene-level image tokenization, (2) a region proposer for discovering regions of interest, (3) a region encoder for region-level image tokenization, and (4) a large language model for modeling multimodal input and output. We detail each component in the following paragraphs.

\subsubsection{Image Encoder.}
Groma employs a pretrained DINOv2~\cite{oquab2023dinov2} model as the image encoder with the input image resolution set to $448$$\times$$448$. Compared with the commonly adopted CLIP~\cite{radford2021learning} visual encoder, DINOv2 is preferred in this work for its compatibility with high-resolution inputs and fine-grained features for localization\footnote{A performance comparison between CLIP and DINOv2 on the detection benchmark is available in our ablation study.}. However, the use of higher-resolution images leads to extended sequences of visual input for the language model, \eg, 1024 tokens in this case. To save computations, we further concatenate every four neighbor patch tokens into a single token following MiniGPT-v2~\cite{chen2023minigpt}. But slightly different from~\cite{chen2023minigpt}, we merge tokens adjacent in 2D instead of 1D, which yields better results empirically. 

\vspace{-3mm}
\subsubsection{Region Proposer.}
To obtain localized understanding of the image, Groma innovatively incorporates a region proposer into the image tokenization process. Specifically, the region proposer is implemented as a class-agnostic detector head using the Deformable DETR (DDETR) transformer~\cite{zhu2020deformable}. The original classification head of DDETR is replaced by a binary classifier to score region proposals based on their localization quality. Inspired by ViTDet~\cite{li2022exploring}, we extract feature maps from the last 4 layers of the image encoder, and rescale these feature maps to construct a hierarchical feature pyramid as the input to the region proposer. For each image, the region proposer generates 300 region proposals, which are then filtered by NMS and objectness scores before fed into the region encoder.

\vspace{-3mm}
\subsubsection{Region Encoder.}
The region encoder translates region proposals (\ie, bounding boxes), coming from both user input and the region proposer, into region tokens. Akin to the previous step, we select feature maps from the last three layers of the image encoder to create a hierarchical feature pyramid. A multi-scale ROIAlign~\cite{he2017mask} module as implemented in \cite{zhang2023gpt4roi, rasheed2023glamm} is utilized to crop and fuse these hierarchical features into unified region tokens. Compared with alternative ways to represent regional inputs, such as numerical representation of positions~\cite{chen2023shikra} and discrete location tokens~\cite{peng2023kosmos, chen2023minigpt}, the region token representation offers distinct benefits as it is semantically aligned with the underlying region, which renders it more intuitive for the language model to comprehend.

\subsubsection{LLM.} We adopt pretrained Vicuna~\cite{chiang2023vicuna} as the language model of Groma. In particular, we instantiate Groma with the 7B version of Vicuna. Besides, we follow LLaVA v1.5~\cite{liu2023improved} to use an MLP layer to project the image tokens and region tokens into the feature space of the LLM.


\subsection{Input and Output Formatting}
\label{sec:format}
Beyond textual only instructions and responses, Groma offers the flexibility to accept user-specified regions as input (referring) and generate visually grounded answers (grounding). Specifically, although different in task formulations, both referring and grounding are unified into one format with the use of region tokens.

\subsubsection{Grounding Output.} 
Remember in the tokenization process, each region token is inherently anchored to a concrete location in the image, corresponding to its region proposal. This connection allows the language model to ground its text output to particular regions in the image by simply referring to the associated region tokens. However, as region tokens are continuous embeddings, they cannot be directly integrated into the codebook of the language model and referenced in the text output. To bridge the gap, we further introduce a set of proxy tokens ``\texttt{<r$_1$>}, \texttt{<r$_2$>}, \texttt{\ldots}, \texttt{<r$_{n}$>}'' to register region tokens. As illustrated below, Groma can refer to any region in the output by addressing the proxy token.

\vspace{2mm}
\noindent
\begin{minipage}{\textwidth}
\begin{tcolorbox} 
\small
\textbf{User}: Here is an image with region crops from it. Image: \texttt{<image>}. Regions: \texttt{<r$_1$>}\texttt{<region>}, \texttt{<r$_2$>}\texttt{<region>}, \texttt{\ldots}, \texttt{<r$_{n}$>}\texttt{<region>}. \texttt{[grounding]} Please briefly describe the image content.

\vspace{1mm}
\textbf{Groma}: \texttt{<p>}A dog\texttt{</p>} \texttt{<roi><r$_4$></roi>} is jumping to catch \texttt{<p>}a frisbee\texttt{</p>} \texttt{<roi><r$_7$></roi>} over \texttt{<p>}a fallen man\texttt{</p>} \texttt{<roi><r$_1$></roi>}.
\end{tcolorbox}
\vspace{1mm}
\end{minipage}
\texttt{<image>} and \texttt{<region>} stand for placeholders of image tokens and region tokens, which are replaced by corresponding visual tokens before being fed into the LLM. \texttt{<p>} and \texttt{</p>} marks the start and end of the grounded phrase. \texttt{<roi>} and \texttt{</roi>} are used to enclose the referenced regions. In addition, we use a special token \texttt{[grounding]} to inform the model to generate grounded responses.

\subsubsection{Referring Input.}
For a region pointed out by the user, we treat it the same as region proposals from the region proposer, \ie, encoding it into a region token and assigning a proxy token to it. This allows us to incorporate user-specified regions into our instructions by inserting corresponding region tokens. A simple example of referential dialogue in Groma is given below, where \texttt{<r$_{10}$>} comes from user-specified region input.

\vspace{2mm}
\noindent
\begin{minipage}{\textwidth}
\begin{tcolorbox} 
\small
\textbf{User}: Here is an image with region crops from it. Image: \texttt{<image>}. Regions: \texttt{<r$_1$>}\texttt{<region>}, \texttt{<r$_2$>}\texttt{<region>}, \texttt{\ldots}, \texttt{<r$_{n}$>}\texttt{<region>}. What is \texttt{<r$_{10}$>}\texttt{<region>}?

\vspace{1mm}
\textbf{Groma}: A cute cat sleeping on a wooden bench.
\end{tcolorbox}
\vspace{1mm}
\end{minipage}

\subsection{Model Training}
\label{sec:train}
The training of Groma is partitioned into three stages: (\textit{i}) detection pretraining for localization ability, (\textit{ii}) alignment pretraining for image-level and region-level vision-language alignment, (\textit{iii}) instruction finetuning for enhanced conversation capability. \cref{table:data} enumerates the datasets used at different training stages. Additionally, we provide the instruction templates used to convert task-specified datasets to instruction following format in Appendix~\ref{sec: templates}.

\begin{table}[h]
    \vspace{-4mm}
    \centering
    \caption{\textbf{Datasets used at three training stages.} RefCOCO/g/+ is short for RefCOCO, RefCOCO+, and RefCOCOg. REC means referring expression comprehension.}
    \vspace{-2mm}
    \label{table:data}
    \tablestyle{8pt}{1.2}
    \resizebox{\linewidth}{!}{
        \fontsize{10pt}{12pt}\selectfont
        \begin{tabular}{l l l}
        \toprule
        Training stage & Data types & Datasets \\
        \midrule
        Detection pretraining & Detection & COCO, Objects365, OpenImages, V3Det, SA1B \\
        \midrule
        \multirow{4}{*}{Alignment pretraining} & Image caption & ShareGPT-4V-PT \\
        & Grounded caption & Flickr30k Entities \\
        & Region caption & Visual Genome, RefCOCOg \\
        & REC & COCO, RefCOCO/g/+, Grit-20m \\
        \midrule
        \multirow{4}{*}{Instruction finetuning} & Grounded caption & Flickr30k Entities \\
        & Region caption & Visual Genome, RefCOCOg \\
        & REC & COCO, RefCOCO/g/+ \\
        & Instruction following & Groma Instruct, LLaVA Instruct, ShareGPT-4V \\
        \bottomrule
        \end{tabular}
    }
    \vspace{-8mm}
\end{table}

\subsubsection{Detection Pretraining.}
This training stage only involves the image encoder and the region proposer, which collectively constitute a DDETR-like detector. The image encoder is kept frozen during training. To endow the region proposer with localization capability, an extensive collection of detection datasets, including COCO~\cite{lin2014microsoft}, Objects365~\cite{shao2019objects365}, OpenImages~\cite{kuznetsova2020open}, and V3Det~\cite{wang2023v3det}, is utilized for large-scale pretraining. Notably, category information is omitted from the training process, with a primary focus on box supervision.

Considering traditional detection data are typically limited to object-level annotations, we complement the training with a two million subset of SA1B~\cite{kirillov2023segment} data filtered by GLEE~\cite{wu2023GLEE}. Original mask annotations of SA1B are transformed into bounding boxes for consistency. The inclusion of this enriched dataset encourages the region proposer to produce region proposals across a wide spectrum of granularities, encompassing not only object instances but also their constituent parts and various background stuff.

\vspace{-2mm}
\subsubsection{Alignment Pretraining.}
To align vision and language feature space of Groma, we pretrain the model on a wide range of vision-language tasks. Specifically, for image-level alignment, we leverage ShareGPT-4V-PT~\cite{chen2023sharegpt4v} for detailed image captioning. For region-level alignment, we engage COCO~\cite{lin2014microsoft}, RefCOCO~\cite{kazemzadeh2014referitgame}, RefCOCO+~\cite{yu2016modeling}, RefCOCOg~\cite{mao2016generation}, and Grit-20m~\cite{peng2023kosmos} for referring expression comprehension (REC), Visual Genome~\cite{krishna2017visual} for region captioning, and Flickr30k Entities~\cite{plummer2015flickr30k} for grounded caption generation. To maintain training efficiency, we focus finetuning efforts on the MLP projection layer and the region encoder, while other modules are kept frozen throughout the training.

\subsubsection{Instruction Finetuning.}
Based on alignment pretraining, we refine the training data to focus exclusively on high-quality datasets and proceed to unfreeze the language model for finetuning purposes. At this stage, LLaVA Instruct~\cite{liu2024visual} and ShareGPT-4V~\cite{chen2023sharegpt4v} are incorporated to improve the conversational and instruction-following capabilities of Groma\footnote{LLaVA Instruct contains three types of instruction data, namely conversation, detailed description, and complex reasoning. Since the detailed description part of LLaVA Instruct has severe hallucinations, we replace it with ShareGPT-4V as in~\cite{chen2023sharegpt4v}.}. Besides, we curate a high-quality grounded chat dataset, named Groma Instruct (see next section for more details), to facilitate synergy of chatting and grounding abilities of Groma.

\vspace{-2mm}
\subsubsection{Discussions.}
A major difference between the training of Groma and current MLLMs is the integration of dedicated detection pretraining, which endows Groma with robust and precise localization ability. Thanks to the decoupled architecture of location and understanding within Groma, we circumvent the need to involve the LLM during detection pretraining. Such a strategic design allows Groma to benefit from pretraining on millions of bounding box annotations — a task that would be computationally prohibitive for classic MLLMs.

\section{GPT4V-assisted Grounded Conversation Generation}
Visual dialogue data have proven to be crucial in advancing the conversational capability of the MLLM as a visual chatbot. Previous methods mostly rely on coarse-grained image descriptions to derive free-form visual dialogues, which typically lack fine-grained region details and precise location information~\cite{liu2024visual, zhu2023minigpt}. For grounded MLLMs, such free-form dialogue data are shown to be insufficient to enable the model to generate long-form grounded responses~\cite{zhang2023llava} - as the format of grounded responses significantly deviates from that of normal responses, it could be challenging for the grounded MLLM to generalize its grounding capability to long-form conversations.

\begin{figure}[h]
\vspace{-6mm}
\centering
\includegraphics[width=1.0 \textwidth]{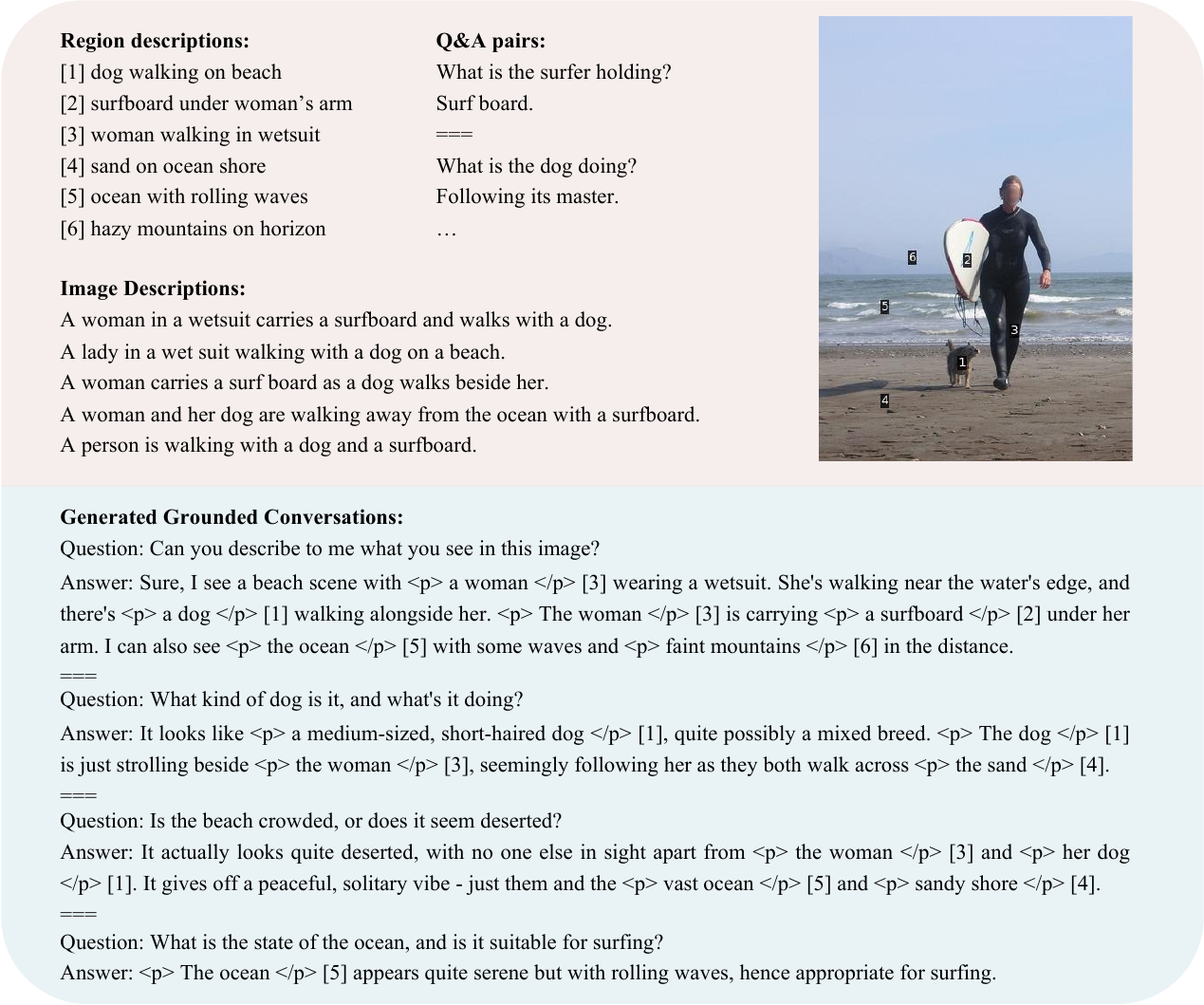}
\caption{\textbf{A data example from Groma Instruct}. The top part demonstrates the image input with visual prompts and contextual text input to GPT-4V. Notably, all numbered regions have corresponding bounding box annotations, which are not shown in the image. The bottom part is the grounded conversations generated by GPT-4V.}
\label{fig:data}
\vspace{-6mm}
\end{figure}

To bridge the gap, we have meticulously curated a dataset containing 30k visually grounded conversations for instruction finetuning, named Groma Instruct. An illustrative example from Groma Instruct is showcased in \cref{fig:data}. Specifically, we select images with dense region annotations from Visual Genome~\cite{krishna2017visual} (VG), and take the following steps to construct grounded conversations with the assistance of advanced GPT-4V model:

\vspace{-1mm}
\begin{itemize}[itemsep=0.5em, leftmargin=*]
\item First, we remove highly overlapped regions (bounding boxes) from VG annotations, normally leaving 3-10 regions of interest for each image. Then we adapt the visual prompting techniques from SoM~\cite{yang2023set} to overlay a bright numeric marker at the center of each region. Using this marked image as input unleashes the grounding capabilities of GPT-4V - it can easily make references to specific image regions by addressing the corresponding numbers. 

\item Besides visual input, we supply GPT-4V with rich region descriptions, image descriptions, and image-based Q\&A pairs, coming from COCO~\cite{lin2014microsoft} and VG annotations\footnote{We select VG images that also have a coco id. Thus, we can retrieve corresponding image captions from COCO Caption.}. While such textual context is optional for GPT-4V input, we empirically find it useful to reduce hallucinations in generated contents and resolve potential ambiguities in visual prompts\footnote{There are cases where two regions highly overlap with each other and GPT-4V can hardly tell from the image which region maps to which numeric marker. For these cases, GPT-4V could rely on the numbered region descriptions to find out correspondences between regions and markers.}.

\item Inspired by prior studies on visual chat data construction~\cite{liu2024visual, zhu2023minigpt, chen2023sharegpt4v, wang2023see}, we further provide GPT-4V with manually designed grounded chat as context examples. This provokes the in-context-learning ability of GPT-4V to generate grounded conversations in a uniform format. We also take a post-processing stage to filter out conversations not following the pre-defined format.
\end{itemize}

\section{Experiments}
In this section, we first quantitatively access the abilities of Groma on grounding (\cref{sec: ground_eval}), referring (\cref{sec: refer_eval}), and image-based conversation (\cref{sec: conv_eval}) tasks. Then we provide qualitative results to exemplify the strong capabilities of Groma on a wide range of region-level tasks (\cref{sec: qualitative}). Finally, we ablate the design and training of Groma in \cref{sec: ablation}.

\subsection{Implementation Details.}
We adopt DINOv2-L/14~\cite{oquab2023dinov2} as the image encoder and Vicuna-7B v1.5~\cite{chiang2023vicuna} as the language model. The region proposer follows an encoder-decoder architecture with 6 encoder layers and 6 decoder layers. We further employ mixed query selection and look-forward-twice scheme as in \cite{zhang2022dino} to accelerate convergence. We set NMS threshold to 0.6 and filter out region proposals with objectness scores lower than 0.15. Subsequently, we select the top 100 region proposals if there are more than 100 proposals left after filtering. This results in no more than 356 visual tokens in total. For training, we sequentially proceed 12 epochs of detection pretraining, 2 epochs of alignment pretraining, and 1 epoch of instruction finetuning. More training details can be found in the Appendix~\ref{sec: more_details}.

\subsection{Grounding Benchmark Results}
\label{sec: ground_eval}

\begin{table}[h]
\vspace{-7mm}
    \centering
    \caption{Results on referring expression comprehension benchmarks. We report accuracy with the IoU threshold set to 0.5. We make Qwen-VL gray because it uses a much larger visual tokenizer (1.9B ViT-bigG~\cite{openclip}).}
    \vspace{-2mm}
    \label{table:ground}
    \tablestyle{4pt}{1.2}
    \resizebox{\linewidth}{!}{
        \begin{tabular}{c c c c c c c c c c c}
        \toprule
        \multirow{2}{*}{Method} & \multirow{2}{*}{Model type} & \multicolumn{3}{c}{RefCOCO} & \multicolumn{3}{c}{RefCOCO+} & \multicolumn{2}{c}{RefCOCOg} & \multirow{2}{*}{Average} \\
        \cmidrule(lr){3-5} \cmidrule(lr){6-8} \cmidrule(lr){9-10}
        & & val & testA & testB & val & testA & testB & val & test & \\
        \midrule
        MDETR~\cite{kamath2021mdetr} & \multirow{3}{*}{Specialist} & 86.75 & 89.58 & 81.41 & 79.52 & 84.09 & 70.62 & 81.64 & 80.89 & 81.81 \\
        G-DINO~\cite{liu2023grounding} & & 90.56 & 93.19 & 88.24 & 82.75 & 88.95 & 75.92 & 86.13 & 87.02 & 86.60 \\
        UNINEXT-L~\cite{yan2023universal} & & 91.43 & 93.73 & 88.93 & 83.09 & 87.90 & 76.15 & 86.91 & 87.48 & 86.95 \\
        \midrule
        VisionLLM~\cite{wang2024visionllm} & \multirow{7}{*}{Generalist} & -- & 86.70 & -- & -- & -- & -- & -- & -- & -- \\
        OFA~\cite{wang2022ofa} & & 79.96 & 83.67 & 76.39 & 68.29 & 76.00 & 61.75 & 67.57 & 67.58 & 72.65 \\
        Shikra~\cite{chen2023shikra} & & 87.01 & 90.61 & 80.24 & 81.60 & 87.36 & 72.12 & 82.27 & 82.19 & 82.93 \\
        Ferret~\cite{you2023ferret} & & 87.49 & 91.35 & 82.45 & 80.78 & 87.38 & 73.14 & 83.93 & 84.76 & 83.91 \\
        MiniGPT-v2~\cite{chen2023minigpt} & & 88.69 & 91.65 & 85.33 & 79.97 & 85.12 & 74.45 & 84.44 & 84.66 & 84.29 \\
        \textcolor{gray}{Qwen-VL}~\cite{bai2023qwen} & & \textcolor{gray}{89.36} & \textcolor{gray}{92.26} & \textcolor{gray}{85.34} & \textcolor{gray}{83.12} & \textcolor{gray}{88.25} & \textcolor{gray}{77.21} & \textcolor{gray}{85.58} & \textcolor{gray}{85.48} & \textcolor{gray}{85.83} \\
        \rowcolor{lightgray}
        Groma & & \textbf{89.53} & \textbf{92.09} & \textbf{86.26} & \textbf{83.90} & \textbf{88.91} & \textbf{78.05} & \textbf{86.37} & \textbf{87.01} & \textbf{86.52} \\
        \bottomrule
        \end{tabular}
    }
\vspace{-5mm}
\end{table}

\noindent
We evaluate the localization capability of Groma on visual grounding tasks.
\cref{table:ground} showcases our performance on three classic referring expression comprehension benchmarks: RefCOCO~\cite{kazemzadeh2014referitgame}, RefCOCO+~\cite{yu2016modeling}, and RefCOCOg~\cite{mao2016generation}. Groma notably surpasses other generalist models of similar model size across all metrics. Even in comparison with Qwen-VL~\cite{bai2023qwen}, which uses a stronger visual tokenizer and trains on $10\times$ more grounding data, Groma delivers superior accuracy on average. Moreover, as a generalist model, Groma shows competitive results with state-of-the-art specialist models~\cite{liu2023grounding, yan2023universal}. These findings underscore the strong capability of Groma in visual grounding.

However, we notice that traditional REC benchmarks only cover a narrow range of common objects in their referring expressions, which is insufficient to thoroughly evaluate the MLLM's localization capability. Therefore, we further introduce LVIS-Ground, an object grounding benchmark converted from the LVIS~\cite{gupta2019lvis} detection data. LVIS-Ground contains 4299 images covering 1203 categories of objects, with on average 3.7 target objects per image. Complementary to REC benchmarks, LVIS-Ground focuses on testing the model's ability to locate multiple, diverse, and variably-sized objects. For more details of LVIS-Ground, please refer to the Appendix~\ref{sec: lvis_ground}.

\cref{table:lvis} presents ours results on LVIS-Ground. Notably, Groma demonstrates clear advantages over other grounded MLLMs, especially on the AR@0.75 metric. This evidences that the specialized design and training indeed bring more accurate localization for Groma. Moreover, it is noteworthy that current MLLMs all fall short of small object localization (AR@s metric). We conjecture this is mainly because the training data (\eg, RefCOCO/g/+, Flickr30k) lack annotations for small objects. We also notice a common failure mode of these methods is that, most of the time they only predict one box per image. This is an expected behavior as the they heavily rely on REC data for grounding training, which only has one target object per query. These findings call for the necessity of diversifying grounding data used for training in future MLLMs.

\begin{table}[ht]
\vspace{-4mm}
    \centering
    \caption{Results on the LVIS-Ground benchmark. We report average recall (AR) to measure performances. For each model, we use the native prompt template recommended by the paper for evaluation.}
    \vspace{-2mm}
    \label{table:lvis}
    \tablestyle{6pt}{1.2}
    \resizebox{0.8\linewidth}{!}{
        \begin{tabular}{ c | c c c | c c c }
        \toprule
        Method & AR & AR@0.5 & AR@0.75 & AR@s & AR@m & AR@l \\
        \midrule
        Shikra~\cite{chen2023shikra} & 4.9 & 14.2 & 2.0 & 0.1 & 3.1 & 18.5 \\
        MiniGPT-v2~\cite{chen2023minigpt} & 11.4 & 19.8 & 11.2 & 0.3 & 8.0 & 41.1 \\
        Ferret~\cite{you2023ferret} & 16.8 & 29.6 & 16.3 & 1.6 & 16.7 & 51.1 \\
        \rowcolor{lightgray}
        Groma & \textbf{28.8} & \textbf{37.9} & \textbf{30.3} & \textbf{8.7} & \textbf{35.6} & \textbf{64.3} \\
        \bottomrule
        \end{tabular}
    }
\vspace{-8mm}
\end{table}

\subsection{Referring Benchmark Results}
\label{sec: refer_eval}

We evaluate Groma on the region captioning task to assess its fine-grained region understanding capability. To prompt the model to generate region-level descriptions, we use queries like \textit{``Please describe <region> in details.''}, where \textit{`<region>'} is replaced by the proxy token and region token corresponding to the target region. \cref{table:refer} presents our results on two established region captioning benchmarks, RefCOCOg and Visual Genome. Without task-specific finetuning, Groma shows comparable or improved performance over GLaMM\footnote{We re-evaluate the performance of GLaMM using the officially released checkpoint after fixing the bug in its original evaluation scripts.}~\cite{rasheed2023glamm}, which has separate designs for input referring and output grounding. This exemplifies the superiority of unified refer-and-ground formulation in Groma.

\begin{table}[ht]
    \centering
    \caption{Results on region captioning benchmarks. We report METEOR and CIDEr scores to measure caption quality. $^\dag$: with task-specific finetuning.}
    \vspace{-2mm}
    \label{table:refer}
    \tablestyle{8pt}{1.2}
    \resizebox{0.7 \linewidth}{!}{
    \begin{tabular}{c c c c c}
    \toprule
    \multirow{2}{*}{Method} & \multicolumn{2}{c}{RefCOCOg} & \multicolumn{2}{c}{Visual Genome} \\
    \cmidrule(lr){2-3} \cmidrule(lr){4-5}
    & METEOR & CIDEr & METEOR & CIDEr \\
    \midrule
    GRIT~\cite{wu2022grit} & 15.2 & 71.6 & 17.1 & 142 \\
    Kosmos-2~\cite{peng2023kosmos} & 14.1 & 62.3 & -- & -- \\
    GPT4RoI~\cite{zhang2023gpt4roi} & -- & -- & 17.4 & 145.2 \\
    GLaMM$^\dag$~\cite{rasheed2023glamm} & 16.1 & 101.9 & \textbf{19.0} & \textbf{163.9} \\
    \rowcolor{lightgray}
    Groma & \textbf{16.8} & \textbf{107.3} & \textbf{19.0} & 158.4 \\
    \bottomrule
    \end{tabular}
    }
\vspace{-2mm}
\end{table}

\subsection{Conversational VQA Benchmark Results}
\label{sec: conv_eval}

In addition to region-level tasks, we further evaluate Groma on the conversational style VQA benchmark, LLaVA Bench (COCO)~\cite{liu2024visual}, which contains three types of questions, namely conversation, detailed description, and complex reasoning. As shown in \cref{table:conv}, Groma surpasses the strong baseline method LLaVA~\cite{liu2024visual} and achieves competitive performance among grounded MLLMs, especially in detailed image description. This demonstrates that Groma maintains decent image understanding and visual chatting abilities. For the underperformance in conversation and complex reasoning questions, we speculate this could be resulted from the DINOv2 features. Recent studies~\cite{jiang2023clip, lin2023sphinx} have shown that DINOv2 image tokenizer slightly underperforms CLIP tokenizer in image understanding tasks as DINOv2 features are not inherently aligned with text. But we believe such gap can be closed by scaling up vision-language alignment pretraining.

\begin{table}[ht]
\vspace{-5mm}
    \centering
    \caption{Results on LLaVA-Bench (COCO).}
    \vspace{-2mm}
    \label{table:conv}
    \tablestyle{8pt}{1.2}
    \resizebox{0.7 \linewidth}{!}{
    \begin{tabular}{c c c c c}
    \toprule
    Method & Conversation & Description & Reasoning & Average \\
    \midrule
    LLaVA~\cite{liu2024visual} & 85.4 & 68.3  & 92.1 & 81.9 \\
    \midrule
    Kosmos-2~\cite{peng2023kosmos} & 71.7 & 63.4 & 74.9 & 70.0 \\
    Shikra~\cite{chen2023shikra} & 80.6 & 70.7 & 88.1 & 79.9 \\
    LLaVA-G~\cite{zhang2023llava} & 79.3 & 71.2 & \underline{92.8} & 81.2 \\
    Ferret~\cite{you2023ferret} & \textbf{84.4} & \underline{79.4} & \textbf{96.3} & \textbf{86.7} \\
    \rowcolor{lightgray}
    Groma & \underline{82.6} & \textbf{84.0} & 88.8 & \underline{85.2} \\
    \bottomrule
    \end{tabular}
    }
\vspace{-8mm}
\end{table}

\subsection{Qualitative Results}
\label{sec: qualitative}

\begin{figure}[h]

\centering
\includegraphics[width=1.0 \textwidth]{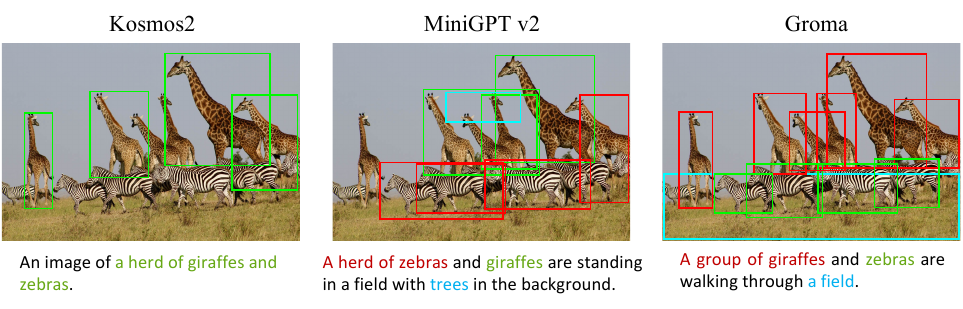}
\vspace{-6mm}
\caption{A qualitative comparison on the grounded image captioning task.}
\label{fig:vis}
\end{figure}

\begin{figure}[h]
\vspace{2mm}
\centering
\includegraphics[width=1.0 \textwidth]{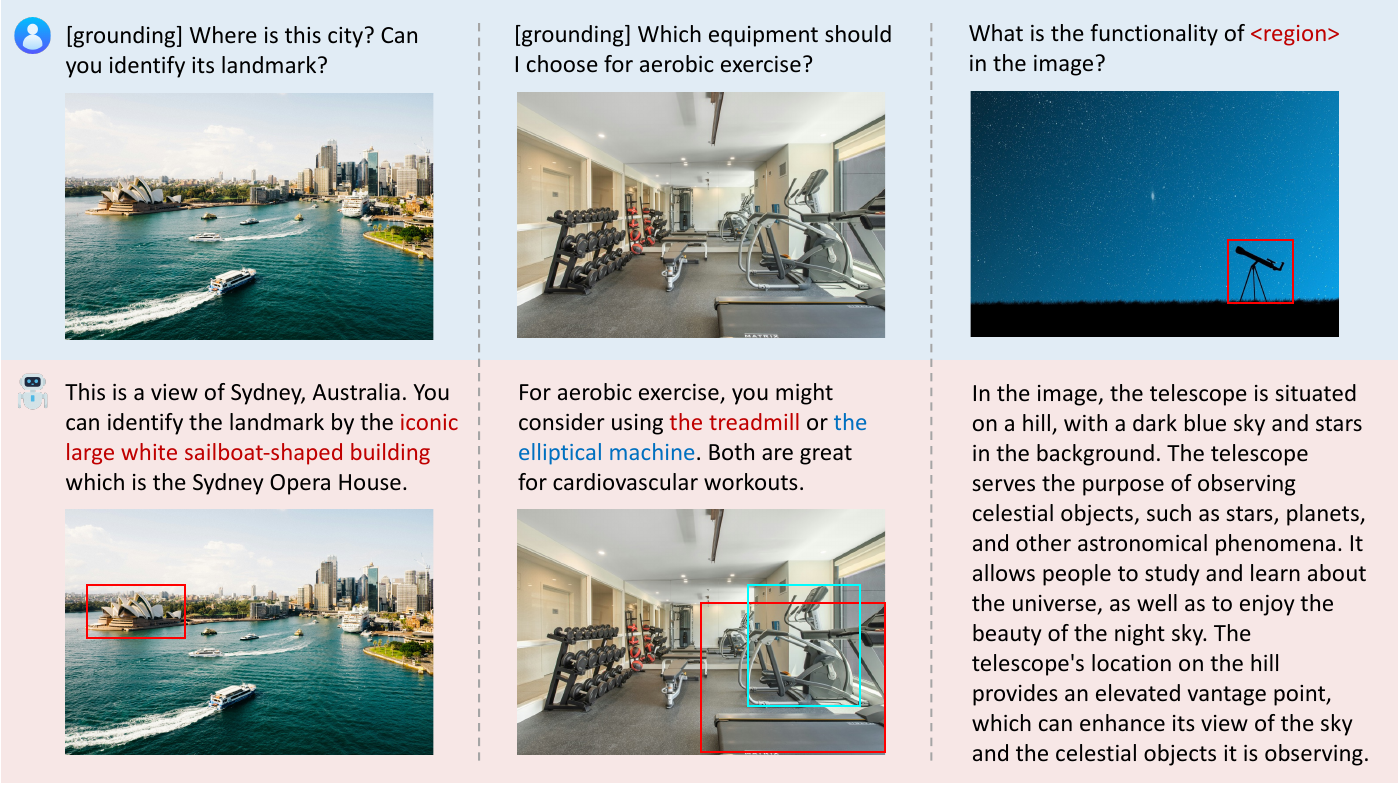}
\caption{Examples on visually grounded chat and referential dialogue.}
\label{fig:demo}
\vspace{-2mm}
\end{figure}

\cref{fig:vis} presents a comparison between Groma and other grounded MLLMs on the grounded image captioning task. We choose an exemplar image that is inherently challenging with multiple and occluded instances to ground. Groma manifests exceptional grounding performance in this case with the highest recall and minimum hallucinations. In addition, we provide several visualization examples in \cref{fig:demo} for a complementary understanding of Groma's abilities on grounded chat and referential dialogue. We show that Groma is capable of generating long-form, grounded and logically rich answers, which can be mainly attributed to the introduction of Groma Instruct data in finetuning. 

\subsection{Ablation}
\label{sec: ablation}

\subsubsection{CLIP \vs DINOv2.} To quantitatively assess the differences in localization capabilities between CLIP and DINOv2, we compare the two backbones on the COCO detection benchmark in \cref{table:backbone}. For this comparison, we equip each backbone with a DDETR~\cite{zhu2020deformable} detection head and finetune only the detection head on COCO dataset. It can be seen that under the same resolution, DINOv2 backbone significantly outperforms CLIP backbone by $6.5$ AP. Furthermore, by scaling the resolution to $448$$\times$$448$, DINOv2 backbone achieves a commendable performance of $43.6$ AP. The results consolidate our choice of DINOv2 backbone in Groma.

\vspace{-2mm}
\subsubsection{Frozen LLM.} In \cref{table:frozen}, we reveal that Groma retains robust localized understanding even without finetuning the LLM, \ie, it demonstrates a referring ability on par with GPT4ROI~\cite{zhang2023gpt4roi} ($148.0$ \vs $145.2$) and grounding ability comparable to Ferret~\cite{you2023ferret} ($84.02\%$ \vs $83.91\%$). This finding suggests our design effectively decouples localization and understanding within Groma, such that it requires minimum `new knowledge' from the LLM for localized understanding.

\subsubsection{Token Merge.} To save computations, Groma by default concatenates every 4 image tokens into one as LLM inputs. Through control experiments in \cref{table:merge}, we find that such downsampling has negligible impacts on the grounding performances (\eg, less than $0.1\%$ average accuracy drop on the REC benchmarks). The results evidence that the decoupled design is optimal in both efficiency and localization accuracy. 

\noindent
\begin{minipage}[t]{0.35\textwidth}
    \centering
    \vspace{-2mm}
    \captionof{table}{Object detection performances on COCO with different backbones and image resolutions.}
    \vspace{1mm}
    \label{table:backbone}
    \tablestyle{6pt}{1.2}
    \fontsize{8pt}{10pt}\selectfont
    \begin{tabularx}{\textwidth}{c c c}
    \toprule
    Backbone & Resolution & AP \\
    \midrule
    CLIP & 336$\times$336 & 32.4 \\
    DINOv2 & 336$\times$336 & 38.9 \\
    DINOv2 & 448$\times$448 & 43.6 \\
    \bottomrule
    \end{tabularx}
\end{minipage}
\hfill
\begin{minipage}[t]{0.6\textwidth}
    \centering
    \vspace{-2mm}
    \captionof{table}{Referring and grounding abilities with the frozen LLM. We measure referring ability with CIDEr score on Visual Genome and grounding ability with average accuracy on REC benchmarks.}
    \vspace{1mm}
    \label{table:frozen}
    \tablestyle{6pt}{1.2}
    \fontsize{8pt}{10pt}\selectfont
    \begin{tabularx}{\textwidth}{c c c c c}
    \toprule
    Stage & LLM status & Referring & Grounding \\
    \midrule
    pretraining & frozen & -- & 82.33 \\
    finetuning & frozen & 148.0 & 84.02 \\
    finetuning & unfrozen & 158.4 & 86.52 \\
    \bottomrule
    \end{tabularx}
\end{minipage}

\begin{table}[h]
\vspace{-2mm}
    \centering
    \caption{Ablation on image token downsampling on the REC benchmarks.}
    \vspace{-2mm}
    \label{table:merge}
    \tablestyle{4pt}{1.2}
    \resizebox{\linewidth}{!}{
        \begin{tabular}{c c c c c c c c c c}
        \toprule
        \multirow{2}{*}{Downsampling} & \multicolumn{3}{c}{RefCOCO} & \multicolumn{3}{c}{RefCOCO+} & \multicolumn{2}{c}{RefCOCOg} & \multirow{2}{*}{Average} \\
        \cmidrule(lr){2-4} \cmidrule(lr){5-7} \cmidrule(lr){8-9}
        & val & testA & testB & val & testA & testB & val & test & \\
        \midrule
        \cmark & 89.32 & 92.15 & 85.96 & 84.11 & 88.10 & 78.40 & 86.33 & 87.40 & 86.47 \\
        \xmark & 89.54 & 92.54 & 86.18 & 83.72 & 88.52 & 78.96 & 86.17 & 86.84 & 86.55 \\
        \bottomrule
        \end{tabular}
    }
\vspace{-4mm}
\end{table}

\section{Limitations and Conclusions}

In this paper, we introduce a novel paradigm, Groma, to unleash the localized perception capabilities of MLLMs. We make the pioneering attempt to embed localization into image tokenization. Our paradigm is based on a perception-then-understand mindset that separates localization from high-level understanding and reasoning. Without introducing external modules, our approach overcomes the resolution bottleneck of using LLMs as location decoders and unifies referring and visual grounding tasks. Extensive experiments showcase the superior performance of our approach in localized perception, as evidenced by its success in referring and visual grounding tasks.

However, the current implementation does not support free-form region inputs and pixel-level grounding. A promising direction to address such limitations is to re-implement the region encoder with a visual sampler as in~\cite{zou2024segment, you2023ferret} and replace the box region proposer by a mask region proposer like Mask2Former~\cite{cheng2022masked}. We leave this for future studies.

\clearpage  

%
%
\bibliographystyle{splncs04}
\bibliography{main}
\clearpage

\appendix

\section{Task-Specified Instruction Templates}
\label{sec: templates}
In complementary to discussion on training datasets in Sec. 3.3, we list a few instruction templates used to convert task-specified datasets to instruction following format in \cref{table:templ}. Specifically, we convert the detection dataset COCO to multiple objects grounding data in a similar format as REC data.

\begin{table}[ht]
    \centering
    \vspace{-2mm}
    \caption{\textbf{Instruction templates.} We randomly select three templates from each task for illustration.}
    \vspace{-2mm}
    \label{table:templ}
    \tablestyle{8pt}{1.2}
    \resizebox{\linewidth}{!}{
    \begin{tabular}{p{3.0cm} | l}
    \toprule
    Task & Template \\
    \midrule
    \multirow{3}{3cm}{Image captioning} 
    & What is this photo about? \\
    & Describe the following image. \\
    & Analyze the image in a comprehensive and detailed manner. \\
    \midrule
    \multirow{3}{3cm}{Region captioning} 
    & What is <region>? \\
    & Please briefly describe <region>. \\
    & Give a concise description of <region>. \\
    \midrule
    \multirow{3}{3cm}{Referring expression comprehension} 
    & Locate <p>\{expression\}</p> in the image. \\
    & Which region matches <p>\{expression\}</p>? \\
    & Identify the region that corresponds to <p>\{expression\}<p>. \\
    \midrule
    \multirow{3}{3cm}{Multiple objects grounding}
    & Locate all <p>\{object class\}</p> in this image. \\
    & Find out all instances of <p>\{object class\}</p> in the image. \\
    & Detect and list each <p>\{object class\}</p> that appears in the picture. \\
    \midrule
    \multirow{3}{3cm}{Grounded image captioning} 
    & [grounding] Give me a short description of the image. \\
    & [grounding] Succinctly summarize what you see in the image. \\
    & [grounding] Please summarize the content of this image in brief. \\
    \midrule
    Grounded chat & [grounding] \{Free-form user instructions\}. \\
    \bottomrule
    \end{tabular}
    }
\vspace{-8mm}
\end{table}

\section{LVIS-Ground Benchmark}
\label{sec: lvis_ground}
Current MLLMs typically do not support detecting multiple categories of objects at the same time. Therefore, to customize the LVIS~\cite{gupta2019lvis} detection benchmark for MLLM evaluation, each time we only select one object class that is included in the image to ground. For instance, the grounding query can be formulated as ``Locate all \{object class name\} in this image''. However, this `one-by-one' evaluation strategy unavoidably leads to low efficiency. To save time and maintain class balance, we randomly sample at most 5 images for each object category\footnote{Some categories have fewer than 5 samples in the original LVIS validation set.} from the LVIS validation set to construct LVIS-Ground.

There are often multiple ground-truth boxes for a query in LVIS-Ground. In such cases, traditional methods either adopt the ANY-Protocol or MERGED-BOXES-Protocol to evaluate performance~\cite{kamath2021mdetr}. To be specific, the ANY-Protocol considers recall to be $100\%$ if the prediction matches any of the ground-truth boxes (\eg, with IoU $>$ 0.5), which fails to truly reflect the model's capability in finding out all object instances. On the other hand, the MERGED-BOXES-Protocol merges all ground-truth boxes into a smallest enclosing box as the ultimate ground-truth box. However, this protocol ignores the atomicity of individual boxes, and is not well-suited for instance-level prediction evaluation.

To better evaluate recall for multiple ground-truths, we propose a new protocol termed AS-MANY-Protocol. This protocol selects the top-k predicted boxes (where k is the number of ground-truth boxes) and measures recall over all ground-truth boxes. For example, if there are 3 out of 5 ground-truth boxes hit by the top-5 predicted boxes, the recall is $60\%$. Besides, we follow common practice in detection~\cite{lin2014microsoft} to calculate average recall over 10 IoU thresholds (ranging from 0.5 to 0.95) as the primary metric on LVIS-Ground. 

\section{More Implementation Details}
\label{sec: more_details}

Table~\ref{table:detail} lists the detailed hyper-parameter configuration used for Groma training. It takes roughly $5/2.5/0.5$ days to finish stage $1/2/3$ training on 8 A100 GPUs. For some large-scale datasets, we merely sample a subset from them during training. The total number of training samples in one epoch is given in~\cref{table:detail}.

\begin{table}[ht]
    \centering
    \vspace{-2mm}
    \caption{\textbf{Training details.} RP, RE, and VLP stand for region proposer, region encoder, and vision-language projector (an MLP), respectively.}
    \vspace{-2mm}
    \label{table:detail}
    \tablestyle{8pt}{1.2}
    \resizebox{\linewidth}{!}{
    \begin{tabular}{c c c c}
    \toprule
    Configuration & Detection pretrain & Alignment pretrain & Instruction finetune \\
    \midrule
    optimizer & AdamW & AdamW & AdamW \\
    epochs & 12 & 2 & 1 \\
    batch size & 64 & 128 & 128 \\
    learning rate & 2e-4 & 1e-4 & 1e-5 \\
    weight decay & 1e-4 & 0 & 0 \\
    resolution & 448p & 448p & 448p \\
    training samples & 5.7m & 3.2m & 857k \\
    trainable param. & RP & RE, VLP & RE, VLP, LLM \\
    \bottomrule
    \end{tabular}
    }
\vspace{-8mm}
\end{table}


\end{document}